# A Deep Learning Pipeline for Solid Waste Detection in Remote Sensing Images

Federico Gibellini [a,*], federico.gibellini@polimi.it
Piero Fraternali [a], piero.fraternali@polimi.it
Giacomo Boracchi [a], giacomo.boracchi@polimi.it
Luca Morandini [a], luca.morandini@polimi.it
Thomas Martinoli [a], thomas.martinoli@polimi.it
Andrea Diecidue [a], andrea.diecidue@polimi.it
Simona Malegori [a], simona.malegori@polimi.it

[a] Department of Electronics, Information, and Bioengineering, Politecnico di Milano, Via Ponzio 34/5, Milan 20133, Italy
[*] Corresponding author

## Abstract

Improper solid waste management represents both a serious threat to ecosystem health and a significant source of revenues for criminal organizations perpetrating environmental crimes. This issue can be mitigated thanks to the increasing availability of Very-High-Resolution Remote Sensing (VHR RS) images. Modern image-analysis tools support automated photo-interpretation and large territory scanning in search of illegal waste disposal sites. This paper illustrates a semi-automatic waste detection pipeline, developed in collaboration with a regional environmental protection agency, for detecting candidate illegal dumping sites in VHR RS images. To optimize the effectiveness of the waste detector at the core of the pipeline, extensive experiments evaluate such design choices as the network architecture, the ground resolution and geographic span of the input images, as well as the pretraining procedures. The best model attains remarkable performance, achieving 92.02 % F1-Score and 94.56 % Accuracy. A generalization study assesses the performance variation when the detector processes images from various territories substantially different from the one used during training, incurring only a moderate performance loss, namely an average 5.1 % decrease in the F1-Score. Finally, an exercise in which expert photo-interpreters compare the effort required to scan large territories with and without support from the waste detector assesses the practical benefit of introducing a computer-aided image analysis tool in a professional environmental protection agency. Results show that a reduction of up to 30 % of the time spent for waste site detection can be attained.

## 1 Introduction

Improper waste handling threatens both ecosystems and human health, causing air, soil, and water pollution [1,2]. Illegal solid waste disposal may alter the dynamics of ecological systems and boost the diffusion of alien or dangerous species, as demonstrated by the correlation between the presence of landfills and the incidence of mosquito-borne illnesses, such as dengue and chikungunya [3]. In addition to health hazards, illegal waste management is among the most fruitful activities for environmental crime organizations. In 2020, the annual revenues for hazardous waste trafficking in the EU were estimated between 1.5 and 1.8 EUR billions, whereas profits from non-hazardous waste trafficking ranged between 1.3 and 10.3 EUR billions [4]. All these threats highlight the need for innovation in the investigation processes of environmental protection agencies and law enforcement authorities against improper solid waste management.

The recent advances in Computer Vision (CV), Deep Learning (DL) and Artificial Intelligence (AI), along with the increased availability of Very-High-Resolution (VHR) Remote Sensing (RS) images, i.e., images whose Ground Sample Distance (GSD) is lower than 50 cm/px, open new opportunities for contrasting this phenomenon. The application of image analysis methods to RS imagery enables automatic scanning of wide regions in search of illegal waste dumping sites and can reduce the need for expensive monitoring infrastructures as well as on-site inspections.

This paper illustrates a semi-automatic pipeline, developed in collaboration with a regional environmental protection agency, for detecting candidate illegal dumping sites in VHR RS images. The pipeline exploits a waste detector implemented by means of a DL-based binary image classifier, which receives as input a VHR RS image and outputs a binary prediction (i.e., "*waste*" / "*no waste*"). To optimize the detector accuracy, extensive experiments, yet widely unexplored in existing literature, compare alternative architectures of the underlying neural network, GSD and geographic spans of the input images, as well as the pretraining procedures. A generalization study assesses the performance variation when the detector processes images from regions significantly different from the one used for model training, a topic rarely addressed in previous works. Finally, a novel exercise in which professional photo-interpreters compare the territory scanning effort with and without the support of the waste detector enables a quantification of the practical benefits of introducing CV techniques for waste detection in VHR RS images in the professionals' everyday working routine.

## 2 Literature review

The ever-increasing quality and availability of RS products, along with the exceptional performance of DL-powered image analysis techniques, have driven interest in various Earth Observation (EO) applications, such as land use and land cover classification [5]. In the domain of waste management, applications include territory analysis for landfill site planning [6] and automated waste detection for identifying and monitoring solid waste dumps from RS images. Over the years, this latter field has witnessed the evolution from manual photo-interpretation of satellite images to the application of the most advanced DL architectures for CV tasks such as image and scene classification, image segmentation and object detection.

**Early works**. Early works appeared when automatic image processing on a large scale was still beyond practical feasibility. RS images were manually photo-interpreted [7], requiring a significant human effort which motivated the shift towards computer-aided solutions. The earliest approaches combined satellite and Geographic Information System (GIS) data [8] or applied traditional (i.e., pre-DL) machine learning and CV methods based on the analysis of spectral or textural features of RS images and on the derivation of spectral indices [9–11].

**DL-based approaches**. Following the success of DL architectures for natural image analysis, researchers started applying DL models to waste detection in RS imagery, addressing tasks such as image and scene classification, pixel-level classification, image segmentation, and object detection. The most relevant approaches addressing these tasks are described in the following paragraphs and summarized in Table 1.

**Image and scene classification methods**. Torres et al. [12] implemented a binary scene classification model based on a ResNet backbone [13] and trained on VHR optical images at different resolutions (with GSD ranging between 20 and 50 cm/px) from the Lombardy region in Italy. The binary waste classifier achieves 94.5% Average Precision (AP) on the 20 cm/px GSD images [12] and 87.99% on the complete data set [14]. Kruse et al. [15] developed a semi-supervised approach based on teacher-student DL architecture for plastic waste detection in Sentinel-2 satellite data. Their method exploits a pixel and a patch classifier, where the former creates a heat map of candidate waste locations and the latter processes the scene in each tile to improve classification accuracy. The pixel classifier achieves 90.46% F1-Score, whereas the patch classifier scores 97.56% in the same metric.

**Pixel-level classification and segmentation methods**. Faizi et al. [16] localized urban waste disposals using a pixel-based Multi-Spectral (MS) image classification approach, whereas [17,18] applied segmentation networks to cluster nearby pixels into objects, which are eventually classified as waste. Yong et al. [19] used a segmentation model based on the DeepLabv3+ architecture [20] to find construction and demolition waste in High Resolution (HR) optical images of urban areas, attaining an F1-Score of 77.4%. Zin et al. [21] trained a classic U-Net binary image segmentation network and applied it to VHR Pléiades Neo images (30 cm/px GSD) over the central region of the West Coast of Malaysia, attaining an Accuracy of 80.6% and deriving a risk exposure map from the prediction maps. The recent study [22] illustrates a two-step approach to detect waste materials in China. The first stage exploits a modified Yolov8 architecture to segment HR (GSD < 2 m/px) 4-band (RGB+NIR) images with respect to three types of materials: industrial solid waste, tailing ponds and other solid waste. Then, a subsequent classification phase operated by a ResNet model discriminates the subclasses of household garbage, construction waste, and mixed garbage. The detector achieves a mean Average Precision (mAP) of 84.1% for the instance segmentation of the five waste categories.

**Object detection methods**. Zhou et al. [23] implemented a solid waste detector (SWDet), based on a modified YOLO network [24] to locate industrial and household waste. The authors tested their approach on a public data set of their conception, called SWAD [23], attaining 77.58% mAP. Sun et al. [25] designed an object detection DL method that exploits a Blocked Channel Attention module to increase the discriminative power of the network and emphasize the relevant features in the input channels. The authors processed satellite images acquired in 2021 over 28 cities worldwide and detected over 1,000 dumpsites. They also performed a generalization study and identified the contextual factors that contribute to the formation of illegal dumping sites. Zhang et al. [26] introduced CascadeDumpNet, a 2-stage DL network for construction and domestic waste detection. In the first stage, an object detection architecture equipped with a novel Contextual Feature Synthesis module optimizes the processing of aerial images. The output of such object detector feeds the second-stage classifier, which receives as input also GIS information and is optimized with automatic hyper-parameter tuning to improve accuracy. The resulting model is applied to MS VHR images (GSD = 0.5 m/px) from Shenzhen, China. The authors compared CascadeDumpNet to other object detectors, demonstrating superior performance by attaining 84.6% mAP. The authors also conducted a generalization study on other Chinese regions and discussed the key environmental factors contributing to the formation of dumpsites. Li et al. [27] built an object detection network exploiting a guidance fusion module, a context awareness module, and a multi-scale interaction module which leverages spatial attention and large kernel convolutions to improve the localization of objects at different scales. The authors tested their detector on two data sets, attaining 58.6% mAP on the publicly available SWAD data set.

*Table 1 Summary and comparison of recent DL approaches for waste detection in RS images.*

| Article | Task | Data set | Image source and resolution | Score |
|---|---|---|---|---|
| Torres et al. [12] | Binary scene classification | ~3000 RGB images | Aerial survey, 0.2 m/px | AP: 94.5%<br>F1-Score: 88.2% |
| Torres et al. [14] | Binary scene classification | 10,434 RGB images | Aerial survey, 0.2 m/px<br>WorldView-3, 0.3 m/px<br>Google Earth, 0.5 m/px | AP: 87.99%<br>F1-Score: 80.70% |
| Kruse et al. [15] | Binary classification | ~4,200,000 pixel spectrograms and 5,407 MS patches | Sentinel-2, 10 m/px | F1-Score: 97.56% |
| Faizi et al. [16] | Pixel-based classification | MS images (number unavailable) | Sentinel-2, 10 m/px<br>Landsat-8, 15 m/px | Accuracy: 94.82%<br>Accuracy: 85% |
| Ulloa-Torrealba et al. [17] | Object-based classification | 1,000 RGB-NIR images | Aerial survey, 0.08 m/px | Overall Accuracy: 80.18% |
| Đidelija et al. [18] | Object-based classification | ~175 RGB images | Pléiades 1B, 2 m/px | Overall Accuracy: 95% |
| Yong et al. [19] | Semantic segmentation | 3,322 RGB images | Esri World Image, 1 m/px | F1-Score: 77.40% |
| Zin et al. [21] | Binary image segmentation | 2,775 RGB images | Pléiades Neo, 0.3 m/px | Accuracy: 80.62% |
| Yu et al. [22] | Instance segmentation | 3,466 RGB-NIR images | Gaofen-1, 2 m/px<br>Gaofen-2, 1 m/px<br>Gaofen-6, 2 m/px | mAP: 84.1% |
| Zhou et al. [23]<br>Li et al. [27] | Object detection<br>Object detection | 1,996 RGB images | WorldView-2 and SPOT, 1.8 m/px | mAP: 77.58%<br>mAP: 58.6% |
| Sun et al. [25]<br>Zhang et al. [26] | Object detection<br>Object detection | 2,219 RGB images | Satellites from Gaofen and SuperView series, 0.3-1 m/px | AP: 70.1%<br>mAP: 84.6% |

**Surveys**. Further details on the topic of solid waste detection in RS images can be found in recent surveys. Papale et al. [28] review the relevant methods based on satellite data for landfill identification with a focus on case studies. Fraternali et al. [29] provide a comprehensive review of the approaches for detecting and monitoring large-scale landfills or urban waste dumps, discussing both methods based on traditional CV techniques and recent DL models. Wang et al. [30] survey waste monitoring methods based on both drone and satellite images and summarize benchmark results from recent studies.

**Data sets.** The AerialWaste data set [14] contains more than 11,700 RS images from 3 different sources and annotated for the tasks of binary scene classification, multi-label image classification and weakly supervised localization. The SWAD data set collects binary-labelled images representing various types of waste materials. It contains 1,996 images at 1.8 m/px resolution acquired in various cities of the Henan Province, China. Finally, the Global Dumpsite Test Data contains the images of four types of dumpsites (domestic, construction, industrial and covered) of the 28 cities analyzed in [25].

### 2.1 Literature gaps

Despite the growing interest in waste detection from RS images, several research topics still require further exploration. Among these are the contribution of different image factors and alternative pretraining procedures to detection performance, the ability of the model to generalize on unseen regions, as well as the practical utility of such models in real-world scenarios.

Given the large size of RS images, most methods divide the original imagery into smaller tiles for processing with waste detection algorithms. The chosen size determines the visible content within each patch, henceforth the *context size*, which affects the ability of the detector to isolate waste materials. The image GSD is another critical parameter: higher resolution improves the capacity to discriminate between different waste materials at the cost of increased acquisition expenses for RS products. Moreover, CV models trained on images with a given GSD may experience reduced performance when applied to imagery acquired at different resolutions. Similarly, the initialization of the DL network represents another design choice potentially impacting the model performance. Traditionally, network weights derived from training on natural images were used to initialize the early layers of the network. More recently, large foundation models specialized for EO tasks rely on backbone networks pretrained on extensive RS image data sets, such as Million-AID [31], SatlasPretrain [32], and the NASA/IBM Harmonized Landsat Sentinel-2 data set, which was released along with the HLS-FM foundation model[1]. These data sets offer a potentially advantageous alternative to pretraining on natural images [33].

Another important research question concerns generalization, i.e., the ability of models trained in one region to produce accurate predictions when applied to territories with different morphology, land cover, land use, and waste distribution.

---

[1] https://www.earthdata.nasa.gov/news/nasa-ibm-openly-release-geospatial-ai-foundation-model-nasa-earth-observation-data

Finally, the current literature lacks public data sets and benchmarks for head-to-head comparison of different methods, as well as studies assessing the practical benefits of automated waste detection within the operational investigation processes of environmental and law enforcement agencies.

This study aims to address these gaps. Our novel contributions include in the first place investigating the impact of image GSD, context size, and different pretraining weights on detection performance. Subsequently, we test the generalization of the best model on various geographical regions. Finally, we assess the practical utility of the detector in the routine activities of professionals.

## 3 Methodology

### 3.1 Overview

The proposed waste detection pipeline, illustrated in Figure 1, receives as input a VHR RS optical image previously acquired by the user or by their agency and covering the entire area of interest. The image is split into geo-referenced squared tiles, which are processed by a binary image classifier that labels them as positive if they contain clues of solid waste and negative otherwise. The classifier returns two outputs: *i)* a numerical score, expressing the model confidence for the input tile to belong to the positive class, and *ii)* a saliency map, computed via Grad-CAM [34], highlighting the image pixels where the classifier focused to deliver a positive prediction. Both outputs can be visualized in any GIS software, overlaid to the input satellite images as shown in Figure 1. Tiles are color-coded on a yellow-to-red scale according to the confidence score output by the classifier. Saliency maps help the user zoom in on the relevant portions of the image to discern the presence of waste materials. This enables the users to filter locations based on the output classification score, thus potentially focusing only on sites detected with high confidence. Using GIS software, they can further integrate additional information about each site, such as population density in the surrounding area or the proximity to protected areas and water bodies. During this inspection phase, environmental agency operators may also assign a risk level to each detected location, combining proprietary information from previous inspections or sensitive legal data that cannot be publicly disclosed. This targeted approach reduces the time otherwise spent on labor-intensive tasks such as visually surveying the entire area of interest, while enabling the operators to shortlist high-priority sites for further action. These shortlisted locations can then be investigated through on-site inspections or drone overflights.

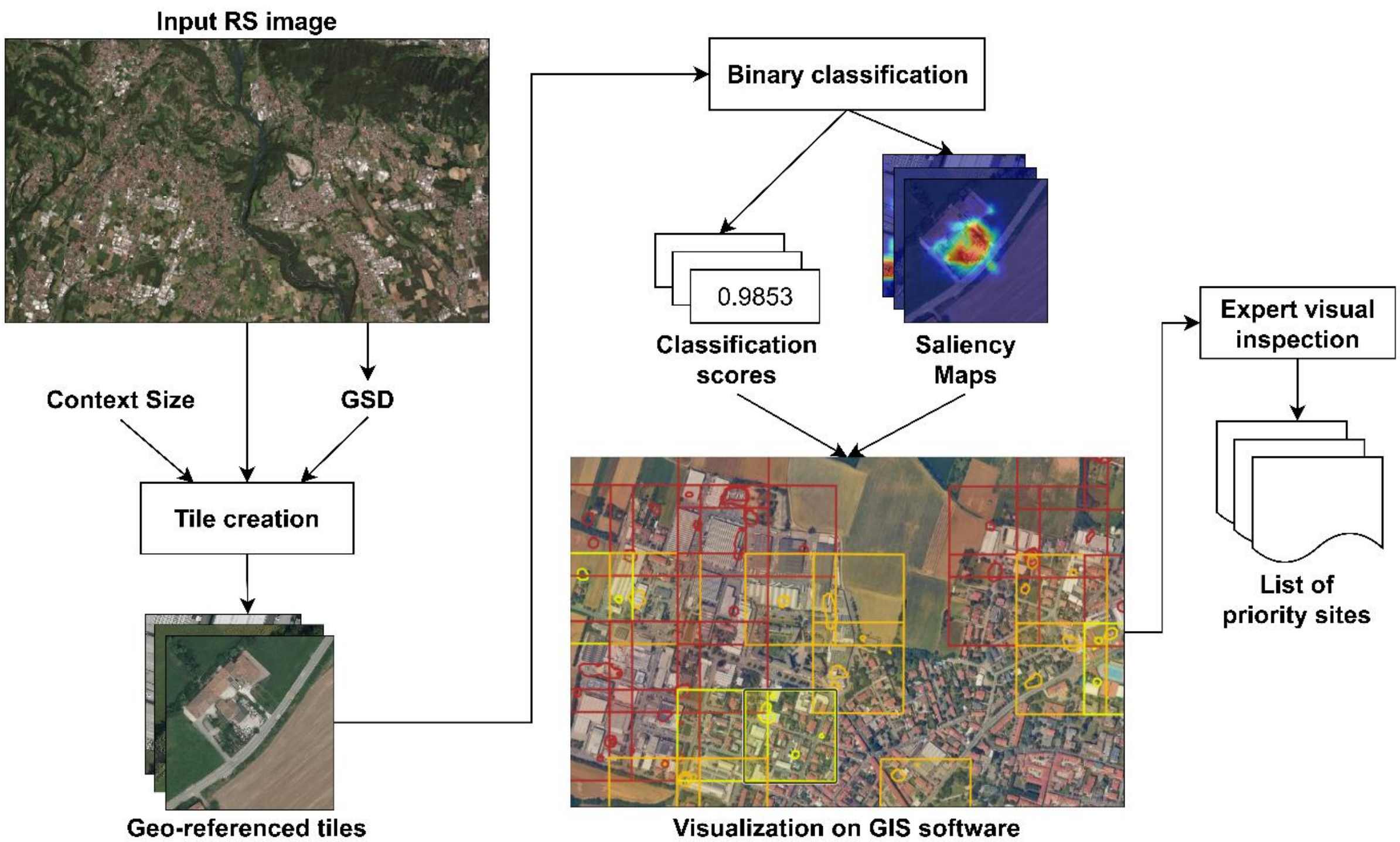

*Figure 1: The RS photo-interpretation process supported by the proposed pipeline.*

### 3.2 Data Set Preparation

The RS image data set used for training and testing the binary waste classifier was created following a location-based approach, thanks to the support from professionals of a local environmental agency. These experts provided the coordinates of waste dumping localities in Lombardy (Italy) known from previous investigations, which determined the positive samples. The locations of negative samples were defined by randomly picking places sufficiently close to the positive sites to guarantee the same geographic context, but also far enough to avoid significant overlaps. During the tile generation step, images were extracted centered in the selected positive and negative locations.

To foster diversity in the RS image data set, tiles were collected from various sources: **Google Earth** (≈21 cm/px GSD in the target area), **WorldView-3** (≈30 cm/px GSD) and **AGEA** (≈20 cm/px GSD). The latter images derive from flights conducted between 2021 and 2023 by the Italian National Agricultural Agency. Professional photo-interpreters visually inspected all the tiles to verify the presence or absence of waste, thus assuring correct labeling. During this revision process, the interpreters ensured that waste appeared in a 100 × 100 m area in the center of the positive samples. This guarantees that cropping an image to reduce its geographic context, as in the experiments described in Section 4.1, does not invalidate the image label. Following these validation phases, the data set consists of ≈11,700 tiles, of which approximately 80% belong to the training set, whereas the remaining belong to the test set. In both sets, the number of negative images is twice the number of positive ones, to account for the real class imbalance between locations with and without waste. This approach is consistent with previous versions of AerialWaste [14], for which this data set was published as Version 3. Figure 2 shows some examples of positive and negative tiles. The presence of waste is highlighted by the saliency maps computed with the best classifier described in the following sections.

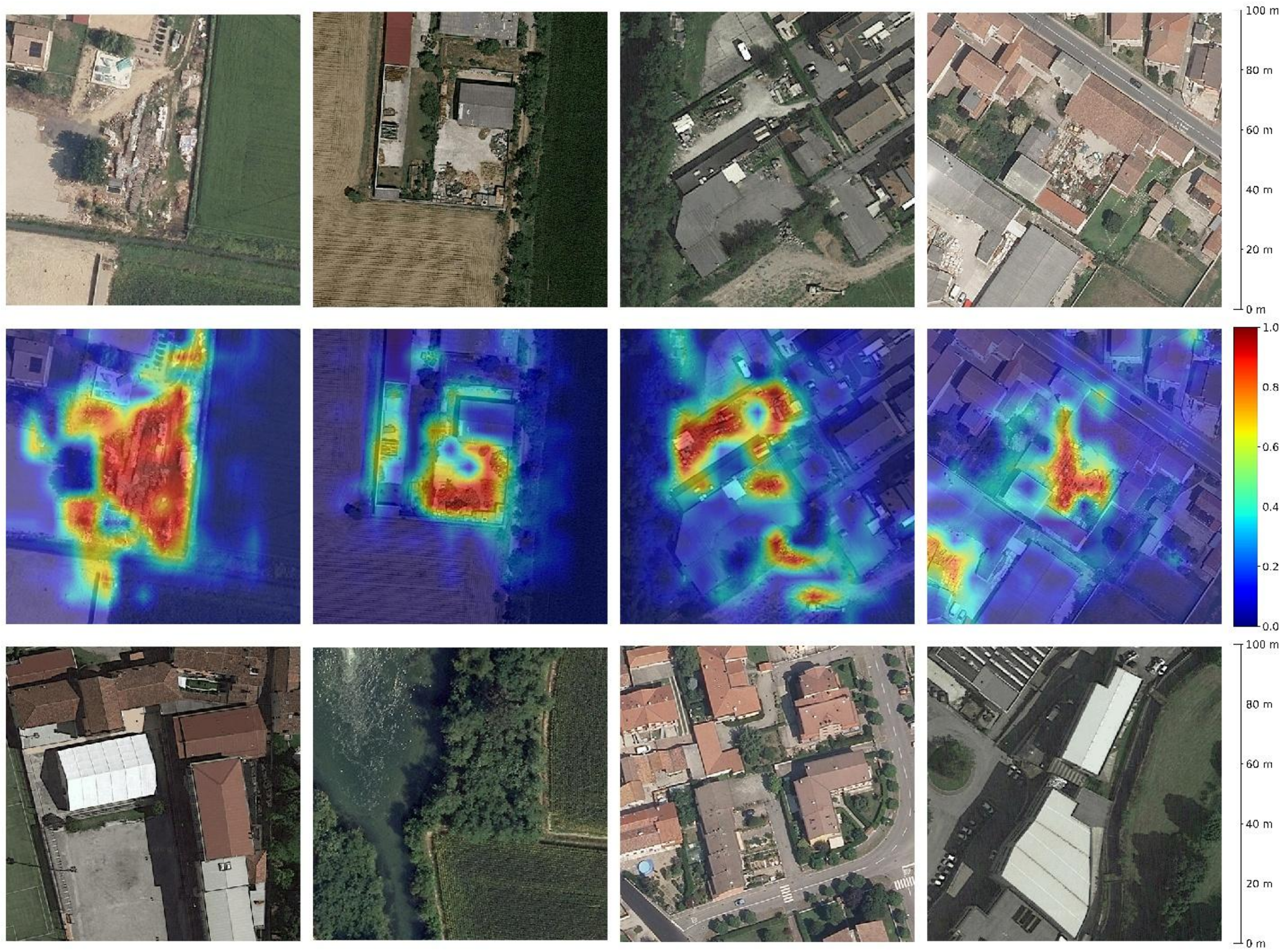

*Figure 2: Examples of positive (first row) and negative (third row) samples from the training data set. The second row highlights the presence of waste (red areas) in positive images by overlaying the saliency maps obtained with the waste detector to the original samples.*

### 3.3 Experimental Setup

The design of the waste detector must account for various factors that affect its performance. These characteristics describe specific features of the input images, such as their GSD and the context size (i.e., the size of the represented geographic area), as well as the detector architecture and its training process.

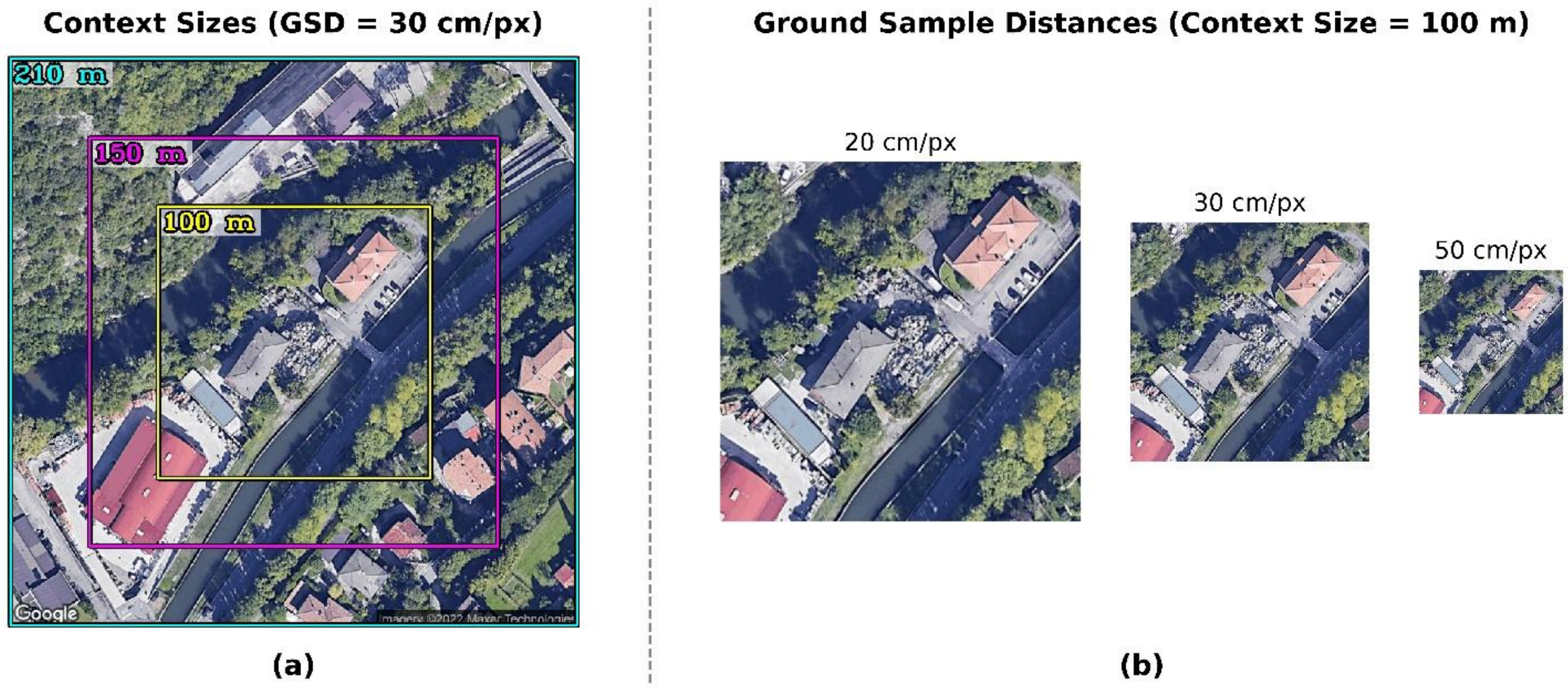

*Figure 3: Effects on the image size of: (a) varying the context size with a fixed GSD; (b) varying the GSD with a fixed context size (100 m). The positive tiles are manually validated to ensure that the smallest context size retains the presence of the waste within the image.*

### 3.3.1 Image Factors: GSD, Context Size, and Image Size

Two factors primarily influence the tile content: *i)* the *GSD*, which denotes the size of one pixel on the ground, expressed in (centi)meters per pixel, and *ii)* the geographic *context size*, which describes the dimensions of the squared geographic area enclosed in the tile, measured in meters. The combination of GSD and context size determines the size in pixels of the resulting tiles. As an example, given a RS image with 30 cm/px GSD, a tile covering an area of 150 × 150 m yields a size of 500 × 500 pixels. In addition, the combination of these two factors affects the detection performance. As illustrated in Figure 3.a, increasing the context size implies including a wider region in a single image, thus providing the network with more information. This may benefit the classification task, because the evaluation of the surrounding context should influence the detection of a specific waste object: for instance, a car in a parking lot should not be considered waste, whereas a car abandoned in a forest should. However, if the GSD is not adjusted accordingly, increasing the context size implies enlarging the image, thus potentially hindering the model classification capabilities. On the other hand, increasing the GSD while keeping the context size fixed (Figure 3.b) produces smaller images with a coarser level of details. This results in losing the distinctive pattern of certain waste objects, such as asbestos plates, whose characteristic corrugated surface may be unnoticeable at a small scale. The chosen GSD values are 20, 30 and 50 cm/px. The 20 cm/px GSD value is the most common one in the training set, whereas the others are offered by commercial VHR satellite products. The selected context size values are 100, 150 and 210 m, with the latter being the default size of the images in the training data set. Smaller values have been defined to reduce the area portrayed by the positive tiles without altering the image label. The resulting image sizes used in the experiments were rounded to the closest multiple of 4 to provide non-padded images as input to the early convolutional layers of the chosen architectures.

### 3.3.2 Training Factors: Network Architectures and Training Procedure

The experiments compare two network architectures, ResNet-50 [13] and Swin-T [35], which are representatives of two popular classes of image classification models, i.e., Convolutional Neural Networks and Vision Transformers. The chosen networks have a comparable number of parameters, 23M for ResNet-50 and 27M for Swin-T, and share a similar four-staged architecture. For the binary classification task, the original network head was replaced by a fully connected layer with a single output neuron, followed by a sigmoid activation to produce the classification score for the positive class.

For the selected architectures, two initializations with different pretraining weights were tested: traditional ImageNet [36] Pretraining (INP) and pretraining on Million-AID, a large data set of RS images for aerial scene recognition [33], henceforth Remote Sensing Pretraining (RSP). Training is conducted following a two-step procedure. In the *Transfer Learning* (TL) phase, the weights of the pretrained model are loaded for the entire network except for the classification head, the backbone is frozen, and the head is trained. In the subsequent *Fine-Tuning* (FT) step, the fourth and final stage in the backbone of the model obtained from TL is unfrozen and trained along with the network head to better adapt high-level features to the waste detection task.

All the combinations of GSD, context size and pretraining weights were used to train both network architectures. All training sessions were executed with a batch size of 120 and with a phase-dependent learning rate: 0.001 during TL and 0.0001 during FT.

## 4 Results and Discussion

Section 4.1 illustrates the results of the tests with a focus on the impact of the factors presented in Section 3.3. Section 4.2 presents a generalization study to assess the performance variation when the waste detector processes images from areas different from the region used for training. Section 4.3 reports the findings of a practical evaluation exercise comparing the photo-interpretation effort of professionals with and without support from the waste detector.

### 4.1 Architecture, image and training

The experiments compare the 36 combinations of network architecture (2), GSD (3), context size (3) and pretraining weights (2), to assess the impact of such factors on classification performance. Table 2 and Figure 4 summarize the results of these experiments. The computation of the Precision, Recall and F1-Score metrics assigns an input tile to the positive class if the output confidence score exceeds 0.5.

*Table 2: Summary of experiment results.*

| Network | GSD [cm/px] | Context Size [m] | Image Size [px] | Metrics (RSP) | | | | Metrics (INP) | |
|---|---|---|---|---|---|---|---|---|---|
| | | | | F1-Score | Precision | Recall | Accuracy | F1-Score | Accuracy |
| ResNet-50 | 20 | 100 | 500 | 90.17% | 91.48% | 88.91% | 93.54% | 86.89% | 91.60% |
| | | 150 | 748 | 89.54% | 91.62% | 87.57% | 93.18% | 85.11% | 90.68% |
| | | 210 | 1048 | 86.95% | **91.96%** | 82.51% | 91.76% | 84.25% | 90.02% |
| | 30 | 100 | 332 | **90.98%** | 91.37% | 90.66% | **94.02%** | 87.53% | **91.88%** |
| | | 150 | 500 | 90.81% | 91.16% | 90.49% | 93.90% | 85.69% | 91.06% |
| | | 210 | 700 | 87.43% | 91.50% | 83.75% | 91.98% | 85.84% | 90.91% |
| | 50 | 100 | 200 | 89.12% | 87.56% | **90.75%** | 92.61% | **87.68%** | 91.86% |
| | | 150 | 300 | 88.97% | 89.95% | 88.03% | 92.73% | 85.62% | 90.77% |
| | | 210 | 420 | 87.53% | 89.95% | 85.29% | 91.91% | 85.47% | 90.47% |
| Swin-T | 20 | 100 | 500 | **92.02%** | 90.02% | **94.13%** | **94.56%** | 90.40% | 93.63% |
| | | 150 | 748 | 91.67% | 90.51% | 92.89% | 94.38% | 89.50% | 93.15% |
| | | 210 | 1048 | 91.47% | **92.17%** | 90.79% | 94.35% | 89.69% | 93.27% |
| | 30 | 100 | 332 | 91.63% | 90.08% | 93.28% | 94.32% | 90.40% | 93.59% |
| | | 150 | 500 | 91.83% | 91.03% | 92.66% | 94.51% | 90.58% | 93.72% |
| | | 210 | 700 | 91.52% | 91.06% | 92.04% | 94.32% | **90.74%** | **93.78%** |
| | 50 | 100 | 200 | 90.98% | 89.10% | 92.98% | 93.86% | 89.57% | 93.01% |
| | | 150 | 300 | 90.61% | 89.87% | 91.44% | 93.69% | 89.86% | 93.17% |
| | | 210 | 420 | 91.48% | 90.57% | 92.41% | 94.26% | 89.83% | 93.20% |

#### 4.1.1 Architectures

Both Swin-T and ResNet-50 networks achieve excellent performance on the waste detection task, consistently averaging around 90% F1-Score. However, for each combination of GSD and context size, Swin-T yields better values than ResNet-50, as shown in Figure 4, where the solid lines representing Swin-T configurations are higher than the dashed lines for ResNet-50 with the same GSD and context size. This behavior may be attributed to the improved ability of transformers to capture global features and relationships, enabling a more comprehensive contextual understanding of the image and, consequently, a more effective evaluation.

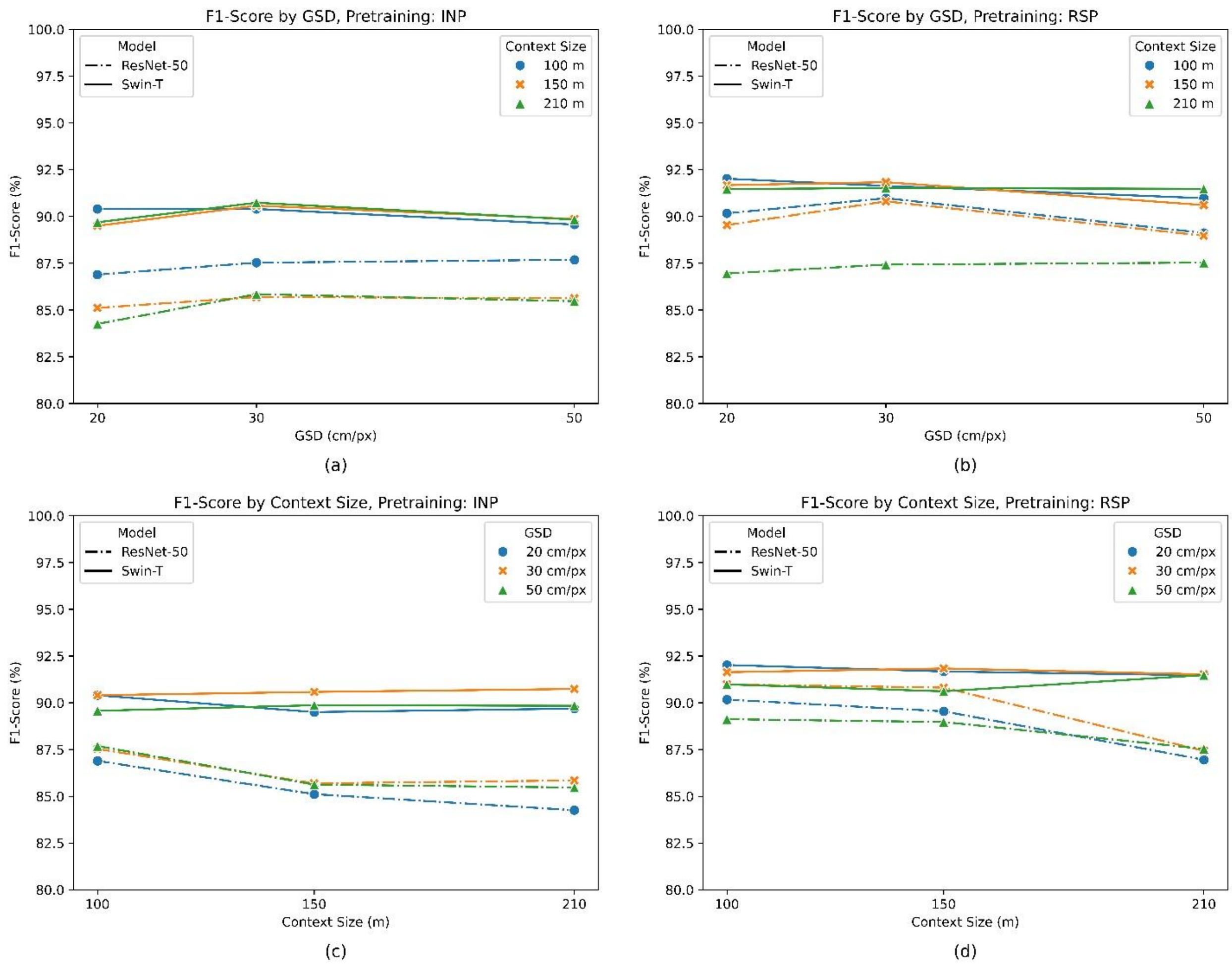

*Figure 4: Distribution of F1-Score across GSD (top) and context size (bottom) values.*

### 4.1.2 Image GSD and context size

The performance of both networks seems not to be influenced by the GSD of the input images, as demonstrated by the horizontal lines in Figure 4.a-b. One possible explanation for this phenomenon is that the determinant features of waste, such as colors and textures, which should guide the model in detection, do not vary significantly across the considered GSD values. These features remain instead recognizable and similar at all tested resolutions, thus resulting in stable performance. Moreover, only a moderate dependency on the context size appears for ResNet-50 models (Figure 4.c-d), with smaller context areas resulting preferable. This behavior might be due to the different nature of the chosen architectures: the convolutional nature of ResNet-50 is accountable for a more significant local dependency, overcome by Swin-T thanks to its attention module, which increases the transformer robustness to variations in image and context size.

### 4.1.3 Training weights

In general, RSP guarantees a minor performance increase to both networks, especially to ResNet-50 models, whose metrics approach those of Swin-T models (Figure 4.b-d). However, such an increase does not suggest a significant improvement in adopting RS images for pretraining rather than natural images (INP). This behavior might be motivated by the different GSD of the images used for backbone pretraining, which do not match those of the waste detection task. Indeed, RSP weights are obtained by training the networks on a very large data set of RS images, whose GSD values vary in the wide range between 0.5 and 153 m/px [31]. Therefore, RSP results in learning features at GSD values higher than 50 cm/px, thus being potentially beneficial when fine-tuning on images with larger GSD values, closer to those of the pretraining images rather than those considered here.

### 4.1.4 Best configuration

In summary, results show that the Swin-T architecture consistently outperforms ResNet-50 across all experiments, thanks to the increased contextual understanding of transformer models. The impact of GSD variations appears negligible, likely due to the absence of significant changes in the characteristic features of waste instances across the considered GSD values. Context size has only a minor influence, which is more evident for ResNet-50 models and tends to favor smaller patches. This behavior can be attributed to the more limited robustness to context increases for networks without attention modules. Finally, pretraining with RSP yields a modest but consistent performance improvement, albeit constrained by the different GSDs of images in the pretraining and fine-tuning data sets.

The best model overall is Swin-T pretrained on RSP and fine-tuned on images with a GSD of 20 cm/px and a context size of 100 m. This model, as shown Table 2, achieves the highest performance in terms of both F1-Score (92.02%) and Accuracy (94.56%).

## 4.2 Generalization study

To assess the classifier generalization capabilities, the model was tested for inference on images from 3 European countries[2]: Greece, Sweden, and Romania. These territories portray landscapes from diverse climate regions as well as differences in visual appearance compared to Lombardy, which provided the images for fine-tuning the waste detector. The Swedish region is the most similar to Lombardy, differing mainly for the significant presence of woods and for the architectural peculiarities of Northern Europe buildings, such as grey roofs. The Romanian region primarily covers an urban landscape where buildings are significantly denser and more cluttered than in Lombardy. Finally, the Greek region covers a hilly territory characterized by dry lands differing from the typical greenery of Lombardy.

[2] The regions were suggested by partner agencies of the PERIVALLON project (https://perivallon-he.eu).

The test sets to assess the model generalization performance were prepared following a location-based approach. Initially, positive and negative sites were manually identified on Google Maps, with care taken to select points that maximize the land-cover heterogeneity, thus portraying urban, rural and industrial areas. Then, at the selected locations, 100 × 100 m squared tiles were extracted from different sources depending on the target country: a WorldView-3 acquisition performed on July 2$^{nd}$ 2024 for the Greek region, Google Earth for the Swedish and Romanian areas. The extracted tiles were visually double-checked to verify the label correctness. This process led to 3 different test sets, whose sample distribution is reported in Table 3. Exemplary samples are presented in Figure 5, Figure 6 and Figure 7.

*Table 3 Distribution of samples across the generalization test sets and metrics scored upon inference with the waste detector.*

| Region | Positives | Negatives | Samples | F1-Score | Precision | Recall | Accuracy |
|---|---|---|---|---|---|---|---|
| Greece | 116 | 137 | 253 | 85.45% | 90.38% | 81.03% | 87.35% |
| Sweden | 187 | 193 | 380 | 83.82% | 91.19% | 77.54% | 85.26% |
| Romania | 185 | 133 | 318 | 91.48% | 96.41% | 87.03% | 90.57% |

The best model identified in Section 4.1.4, i.e., Swin-T pretrained with RSP weights and trained on images with context size of 100 m and GSD of 20 cm/px, was run on such generalization test sets, achieving the results presented in Table 3. Averaging across the three regions, the detector scores 87.73% Accuracy and 86.92% F1-Score, thus losing respectively 6.83% and 5.10% compared to the performance on the Lombardy test set. This demonstrates that the classifier generalizes well to regions which are visually and morphologically different from the training area.

The saliency maps in Figure 5, Figure 6, and Figure 7 demonstrate that the classifier can correctly focus on regions covered by waste, confirming its generalization capabilities. For true positives, the network successfully attends to the correct areas, precisely identifying the locations where waste is present. This effect is particularly evident in the Romanian data set, where the saliency maps appear especially accurate. However, this may be partially due to the prevalence of large landfills in the region, which are easier to detect. In the case of false positives, the network often focuses on genuinely suspicious regions that were annotated as negatives, probably because of their reduced dimensions or for being difficult to classify even for a human interpreter. False positives in Sweden highlight these issues since they mainly portray industrial areas where small amounts of residual materials were left outside of factories. Examples of false negatives show that the network can still focus on potentially relevant regions, even though the classification score is lower than the threshold. The Swedish false negatives typically correspond to small landfills in residential areas whose limited size reduces the overall confidence of the classifier. In Greece waste appears in larger quantities of scattered materials, which visually resemble the rocky surroundings. Finally, regarding true negatives, the model correctly rejects clean areas across a variety of contexts, including both rural and urban environments. In rural contexts, the prevalence of fields with regular and uniform textures enhances the model prediction confidence, whereas the abundant cluttering of urban scenes might hinders it. The tested model correctly evaluates both situations. Nevertheless, further improvements could be achieved by fine-tuning the classifier on a data set including images from different regions with diverse contexts.

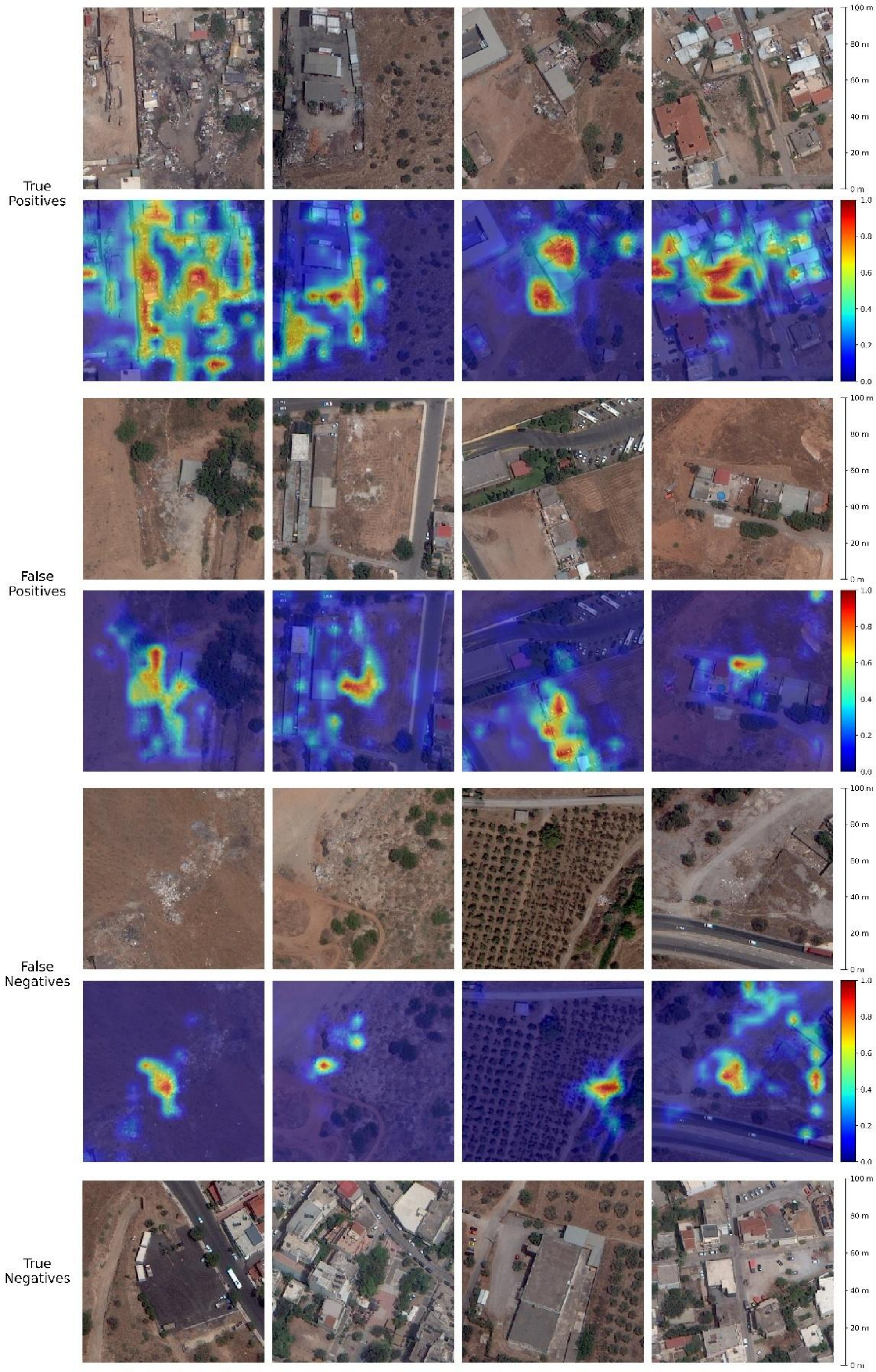

*Figure 5: Examples of true positive, false positive, false negative and true negative samples from the Greek generalization test set [© DigitalGlobe, Inc. (2024), provided by European Space Imaging]. For the first 3 sets of samples, the second row highlights the presence of waste (red areas) by overlaying the saliency maps obtained with the waste detector to the original sample.*

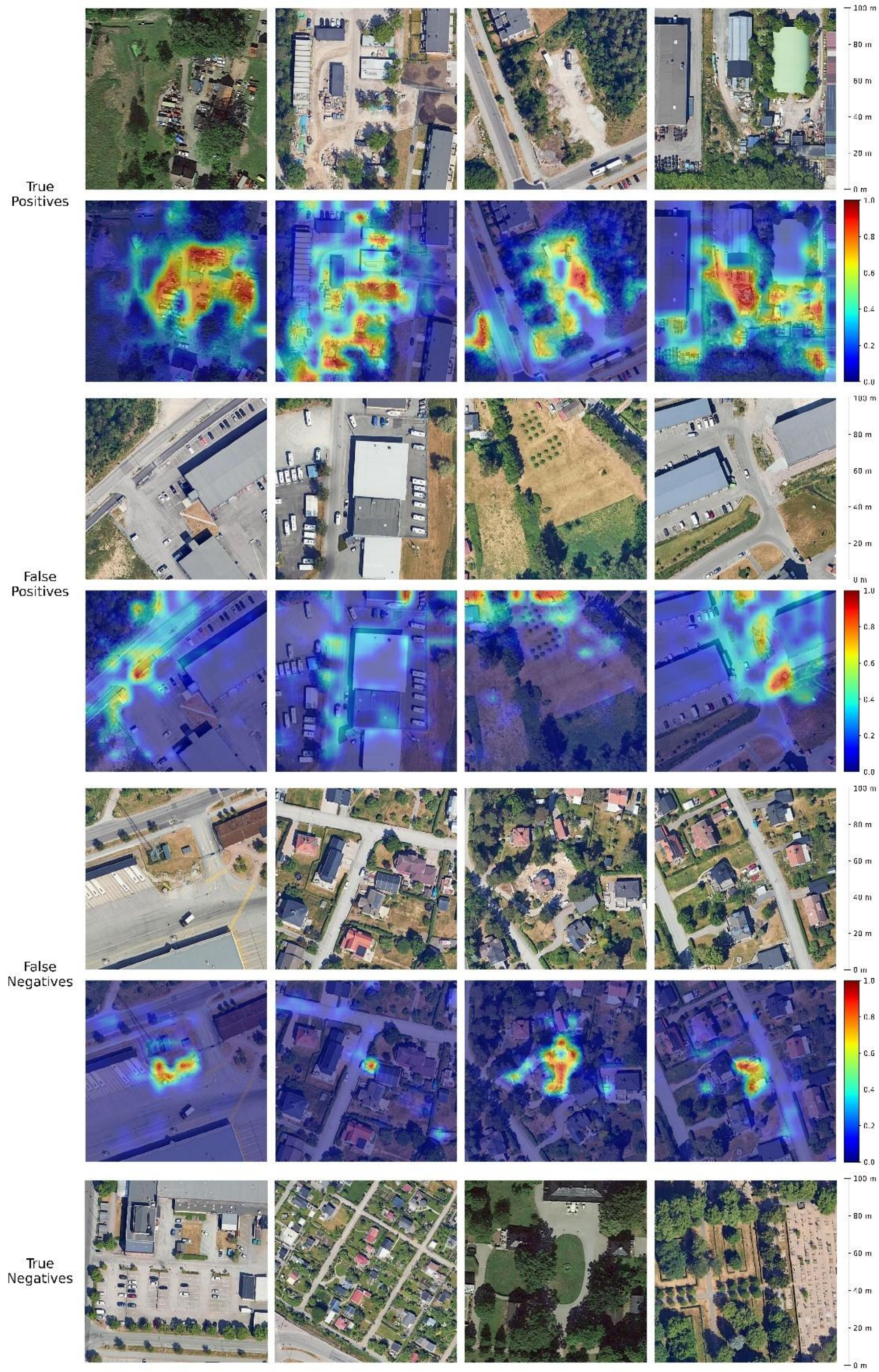

*Figure 6: Examples of true positive, false positive, false negative and true negative samples from the Swedish generalization test set [© Google, 2025]. For the first 3 sets of samples, the second row highlights the presence of waste (red areas) by overlaying the saliency maps obtained with the waste detector to the original sample.*

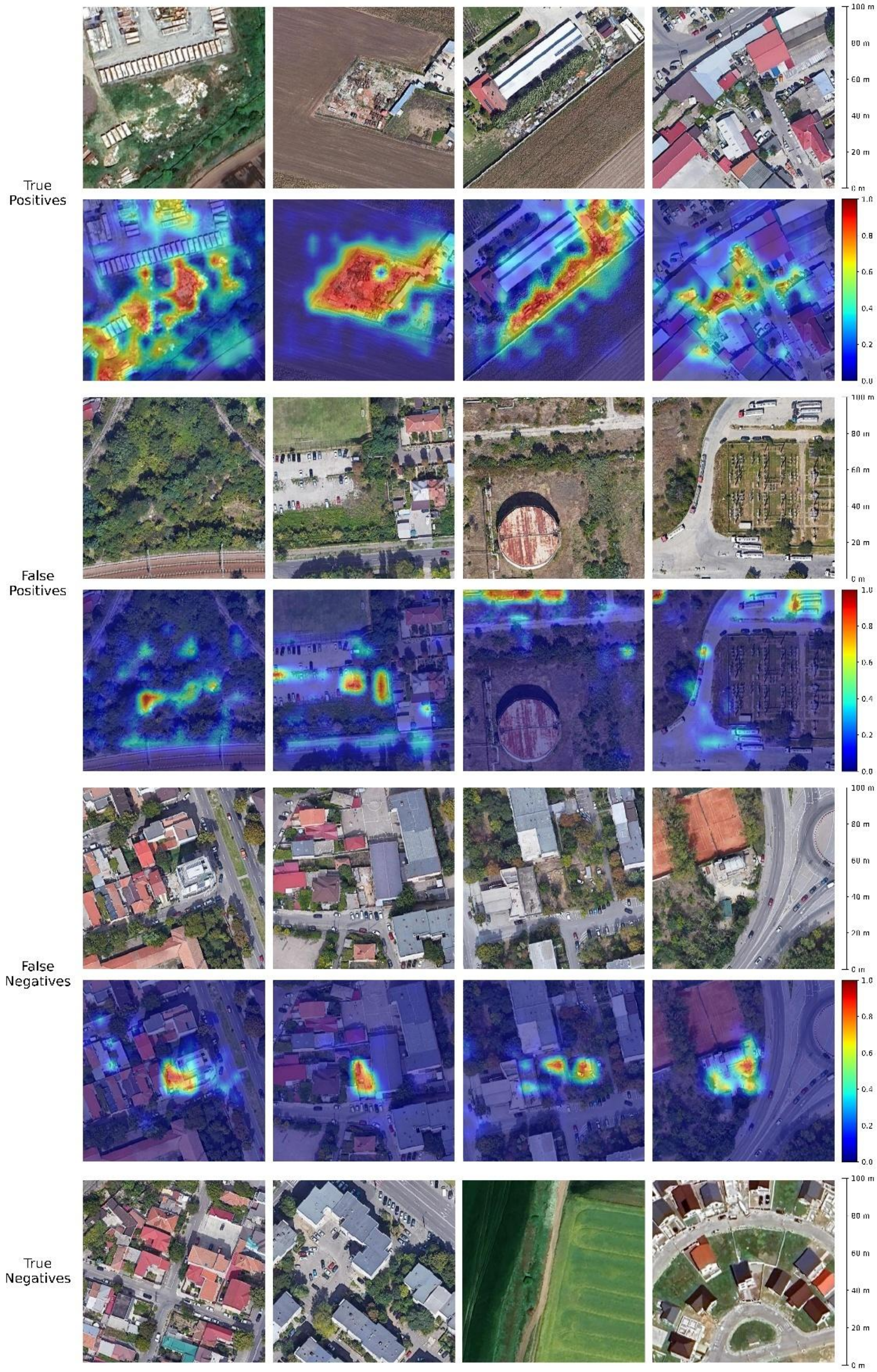

*Figure 7: Examples of true positive, false positive, false negative and true negative samples from the Romanian generalization test set [© Google, 2025]. For the first 3 sets of samples, the second row highlights the presence of waste (red areas) by overlaying the saliency maps obtained with the waste detector to the original sample.*

### 4.3 Utility evaluation

The practical utility of the proposed pipeline was evaluated through a study conducted in collaboration with the same environmental monitoring professionals who provided the database of waste locations to initially build the data set. The goal of such study was to quantify the time saved when scanning a large territory with support from the designed pipeline as opposed to without. The study addressed 12 municipalities in Lombardy, which were processed following two procedures: traditional manual photo-interpretation and an AI-supported approach. The first procedure consisted in manually inspecting the satellite imagery covering the region of interest via a GIS tool, without further assistance. The second method leveraged the proposed pipeline to analyze the entire territory by dividing it into squared tiles and processing them with the waste detector. Then, human interpreters visualized on a GIS tool the scores and saliency maps output by the classifier for all and only the tiles whose classification score was greater than 0.2. Therefore, only the area covered by such tiles was inspected.

The study involved 4 professionals, each of whom analyzed all 12 municipalities, 6 with the traditional method and 6 with support from the classifier. As reported in Table 4, exploiting the classifier predictions allowed the personnel of the environmental agency to detect a larger number of critical sites, 155 against the 95 identified without support from the image-analysis tool, while examining in detail a 60.2% smaller area. Time spent by the professionals to detect and verify each site was recorded to compare the human effort required by each procedure. As expected, the AI-supported procedure yielded a higher overall inspection time due to the larger number of detected sites, but reduced the average time to inspect a site by 12.2%.

Since the experiment aimed at quantifying the human effort in the experiment, time for running the classifier was excluded from computations, as such process does not require any human supervision. Moreover, the time required for model inference on a large area (100 km$^2$) on a single consumer-level NVIDIA GeForce RTX 2080 Ti with 12GB RAM demonstrated to be negligible with respect to the time needed to visually inspect images.

The experiment was then repeated in a more conservative scenario where the classifier threshold was set to 0.7 following the suggestions of the experts involved in the experiment, who reckoned such value to be the best option for detecting all high-risk sites while minimizing the inspection effort. By raising the threshold to 0.7, only tiles with a high confidence of containing waste sites were reported for inspection, thus further and significantly reducing the area to be examined by 64% with respect to the case using a 0.2 filtering threshold. Under these conditions, the number of detected sites also decreased by 32%, as several less confident model detections are filtered out due to the higher threshold. Consequently, inspection with AI support resulted in an approximate 30% time reduction compared with the fully manual photo-interpretation approach. On top of this, the limited number of municipalities addressed in the study did not allow to assess the impact of fatigue on the operators, a relevant phenomenon occurring when analyzing a significant amount of images. Therefore, the measured time savings must be considered a lower bound to the potential gain achieved when analyzing hundreds of municipalities.

*Table 4: Results from the experiment involving professionals from a local environmental agency to evaluate time savings implied by the introduction of the proposed pipeline.*

| Photo-interpretation approach | Manually inspected area [km$^2$] | Detected sites | Total time [min] | Average time per site [min] |
|---|---|---|---|---|
| Without AI support | 125.24 | 95 | 1486 | 15.6 |
| With AI support | 49.84 | 155 | 2133 | 13.7 |
| Variation | -60.2% | **+63.2%** | +43.5% | **-12.2%** |

## 5 Conclusions

Given the threats posed by improper waste management from private citizens and criminal organizations, the application of CV techniques to the RS image analysis can help environmental and law enforcement agencies to increase the frequency and scale of their territory monitoring activities, thus improving the effectiveness and efficiency of their investigation processes. This paper aims to contribute to a better understanding of the decisions involved in the design of a waste detection pipeline, of the generalization capacity of CV models and of the practical utility of applying AI-based tools in the RS image interpretation process. Towards such a goal, the paper provided the following contributions.

A CV pipeline is developed to support environmental and law enforcement agencies in their investigation processes. The pipeline receives as input RS imagery over a region of interest, divides it into tiles on a regular grid pattern, processes such tiles with a DL binary waste classifier and outputs for each processed tile a confidence score and a saliency map. The former denotes the trust of the model in the presence of waste materials in the tile, the latter highlights the regions of the image where waste materials have been detected. The pipeline can be easily integrated by environmental monitoring professionals in their routine workflows. The only potential barrier towards a more systematic adoption is represented by the high costs for acquiring VHR RS imagery, which are however expected to decrease in the near future.

An empirical study assesses how the performance of the binary waste classifier is affected by various factors: *i)* the network architecture, *ii)* the image GSD, *iii)* the size of the area portrayed in the picture, and *iv)* the weights adopted for pretraining the network. The experiments highlight that the best configuration uses a Swin-T model initialized with RSP weights and trained with images at 20 cm/px GSD portraying a squared area with a side of 100 m. Such model achieves 94.56% Accuracy and 92.02% F1-Score. All experiments employ a data set developed in collaboration with expert photo-interpreters from an environmental agency and made publicly available as Version 3 of AerialWaste [12,14].

A generalization study shows that the designed approach retains good performance when applied to RS images of territories with different visual appearance from the training region, averaging 87.73% Accuracy and 86.92% F1-Score. In addition, an evaluation exercise quantifies the impact of the proposed pipeline in the working routine of professionals from an environmental agency. In comparison to manual photo-interpretation, the use of the binary waste detector during the image search saves ≈ 12% time in the analysis of the region of interest and enables the detection of ≈ 63% more candidate waste dumping sites.

This study adds to the very few that perform a generalization test on territories with very distinct characteristics and, to the best of our knowledge, it is the first documented exercise evaluating the practical utility of computer support to photo-interpretation for waste detection. The developed binary waste classifier is being used to locate waste dumping sites in various European countries and its code is publicly available at https://github.com/gblfrc/waste-detection-dl-pipeline.

Future work will focus on extending the training data set with images from various regions to improve its generalization abilities and on developing methods to support prioritization of on-site inspection campaigns. These methods include automatically identifying the specific materials contained in a waste dump and highlighting the presence of particularly hazardous substances such as asbestos or flammable items. Material recognition requires a more complex detector architecture and possibly the use of multi- or hyper-spectral data, to discriminate materials with a similar visual appearance but different spectrographic response, e.g., asbestos plates against corrugated metal sheets. Risk assessment could exploit multi-modal information as input, such as cadaster information, land cover / land use maps and other GIS data. Finally, multi-temporal analysis based on time series of RS images of the same region could be applied to prioritize the investigation of dumping sites with a growing evolution profile.

**Acknowledgements:** Special thanks to the professional photo-interpreters from ARPA Lombardia, who participated in the data set creation process and in the utility validation experiment. This work was supported by the Ministry of Education, Youth and Sports of the Czech Republic through the e-INFRA CZ (ID:90254).

**Author contributions:** Conceptualization, P.F., G.B., F.G., L.M. and T.M.; methodology, P.F. and G.B.; software, F.G. and S.M.; data curation, A.D., F.G., L.M., T.M. and S.M.; writing— original draft preparation P.F. and F.G.; writing—review and editing, F.G., P.F., G.B., L.M., T.M., A.D. and S.M.; visualization, F.G.; supervision, P.F. and G.B. All authors have read and agreed to the published version of the manuscript.

**Funding:** This research was partially funded by European Union's Horizon Europe project PERIVALLON - Protecting the EuRopean terrItory from organised enVironmentAl crime through inteLLigent threat detectiON tools, under grant agreement no. 101073952.

**Data availability statement:** The AerialWaste data set employed in the study is openly available on Zenodo at https://doi.org/10.5281/zenodo.12607190.

**Conflicts of interest:** The authors declare no conflicts of interest.